\documentclass[letterpaper, 10 pt, conference]{ieeeconf}  

\IEEEoverridecommandlockouts                              

\usepackage{graphics} 
\usepackage{epsfig} 
\usepackage{times} 
\usepackage{amsmath} 
\usepackage{amssymb}  
\usepackage{subfigure}
\usepackage{multirow}
\usepackage{algorithm}
\usepackage{algpseudocode}
\usepackage{hyperref}

\makeatletter

\renewcommand{\maketag@@@}[1]{\hbox{\m@th\normalsize\normalfont#1}}%

\makeatother

\title{\LARGE \bf
Towards Autonomous Indoor Parking: A Globally Consistent Semantic SLAM System and A Semantic Localization Subsystem
}

\author{Yichen Sha\textsuperscript{\dag1}, Siting Zhu\textsuperscript{\dag1}, Hekui Guo\textsuperscript{2}, Zhong Wang\textsuperscript{1}, and Hesheng Wang\textsuperscript{1}
\thanks{\textsuperscript{\dag}The first two authors contributed equally.}
\thanks{*This work was supported in part by the Natural Science Foundation of China under Grant 62225309, U24A20278, 62361166632, U21A20480 and 62403311. Corresponding Author: Hesheng Wang ({\tt\small wanghesheng@sjtu.edu.cn}).}
\thanks{$^{1}$School of Automation and Intelligent Sensing, Key Laboratory of System Control and Information Processing, Ministry of Education of China, Shanghai Jiao Tong University, Shanghai.}%
\thanks{$^{2}$Dimensional Robot, Shanghai, China.}%
}

\begin{document}

\maketitle
\thispagestyle{empty}
\pagestyle{empty}

\begin{abstract}

We propose a globally consistent semantic SLAM system (GCSLAM) and a semantic-fusion localization subsystem (SF-Loc), which achieves accurate semantic mapping and robust localization in complex parking lots. Visual cameras (front-view and surround-view), IMU, and wheel encoder form the input sensor configuration of our system. The first part of our work is GCSLAM. GCSLAM introduces a semantic-constrained factor graph for the optimization of poses and semantic map, which incorporates innovative error terms based on multi-sensor data and BEV (bird's-eye view) semantic information. Additionally, GCSLAM integrates a Global Slot Management module that stores  and manages parking slot observations. SF-Loc is the second part of our work, which leverages the semantic map built by GCSLAM to conduct map-based localization. SF-Loc integrates registration results and odometry poses with a novel factor graph. Our system demonstrates superior performance over existing SLAM on two real-world datasets, showing excellent capabilities in robust global localization and precise semantic mapping.

\end{abstract}

\section{Introduction}

Simultaneous Localization and Mapping (SLAM) is a fundamental task in robotics and autonomous driving. Although SLAM has been extensively researched and applied in various scenarios, its application in indoor parking environments presents unique challenges that remain unsolved. In indoor parking lots, the absence of GNSS signals makes it impossible to directly obtain high-precision vehicle poses. Moreover, low-cost sensors such as IMUs and cameras are preferred in commercial applications. 
These sensors face challenges in indoor parking environments, due to the repeated structural features and similar textures across different regions, which hinder the effectiveness of visual SLAM. 

Recently, some methods~\cite{qin2020avp, shao2020tightly, shao2021mofis, zhao2019visual} have shown capability in performing SLAM within indoor parking environments for autonomous valet parking (AVP) tasks.
However, these SLAM systems fail to establish adequate constraints for optimization in environments with highly repetitive structure, which leads to poor performance in large and intricate parking lots.
Additionally, test scenes of these SLAM methods are relatively simple, typically involving only a few rows of parking slots. However, real-world parking lots are complicated, characterized by a high density of parking slots and usually cover large areas that require long-distance driving within the lot. In such scenes, existing methods suffer from accumulative drift in tracking, leading to misalignment and distortion in mapping.
Furthermore, these methods fail to handle noise and false detections in semantic segmentation, leading to inaccurate mapping results and decreased localization accuracy.

\begin{figure}[t]
    \centering
    \includegraphics[width=8.8cm]{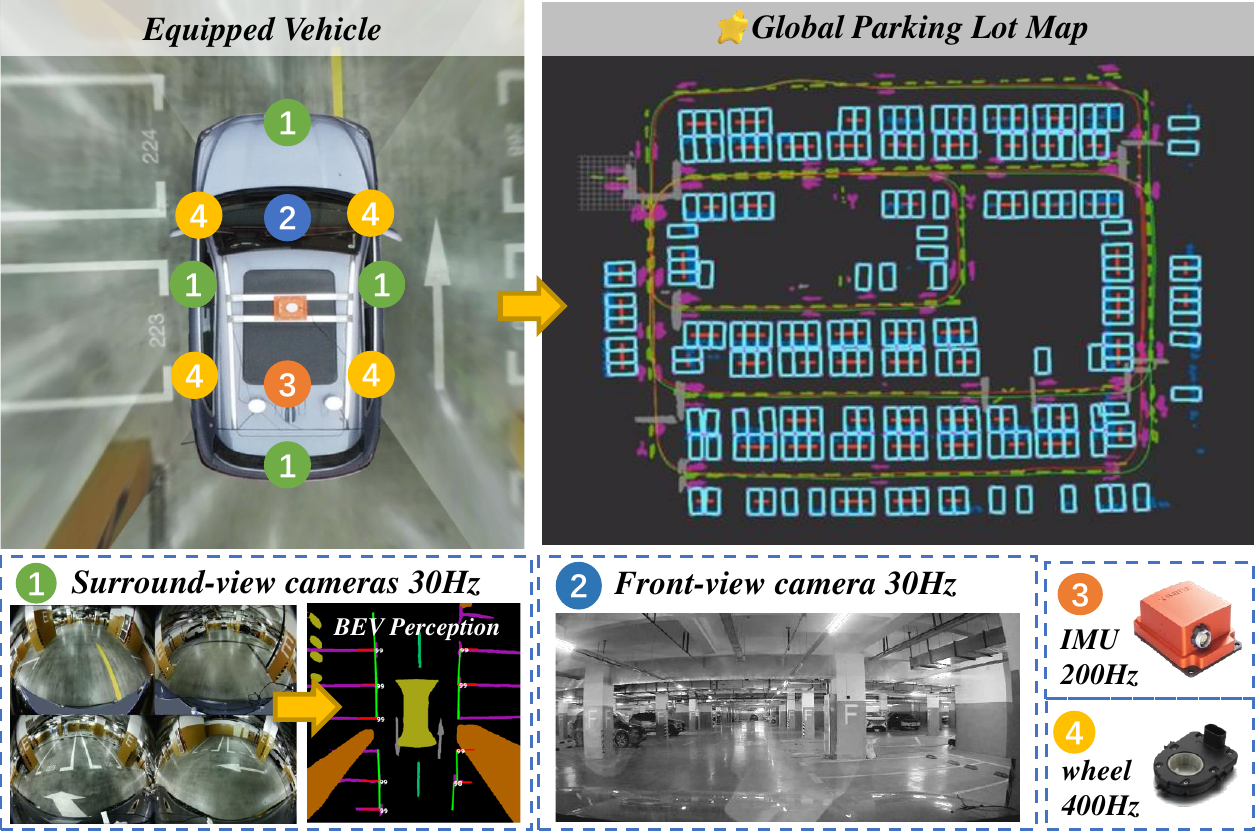}
    \caption{Sensor setup of our system. Our test vehicle is equipped with a front-view camera, an IMU, a wheel encoder, and a surround-view camera system comprising four fisheye cameras. The surround-view images are transformed into a BEV image and processed through a BEV perception module to obtain semantic information. By utilizing these sensor data and features, our system is capable of performing high-precision localization and building a globally consistent semantic map.}
    \label{fig:overal}
    \vspace{-0.2in}
\end{figure}

\begin{figure*}[!htp]
    \centering
    \includegraphics[width=18cm]{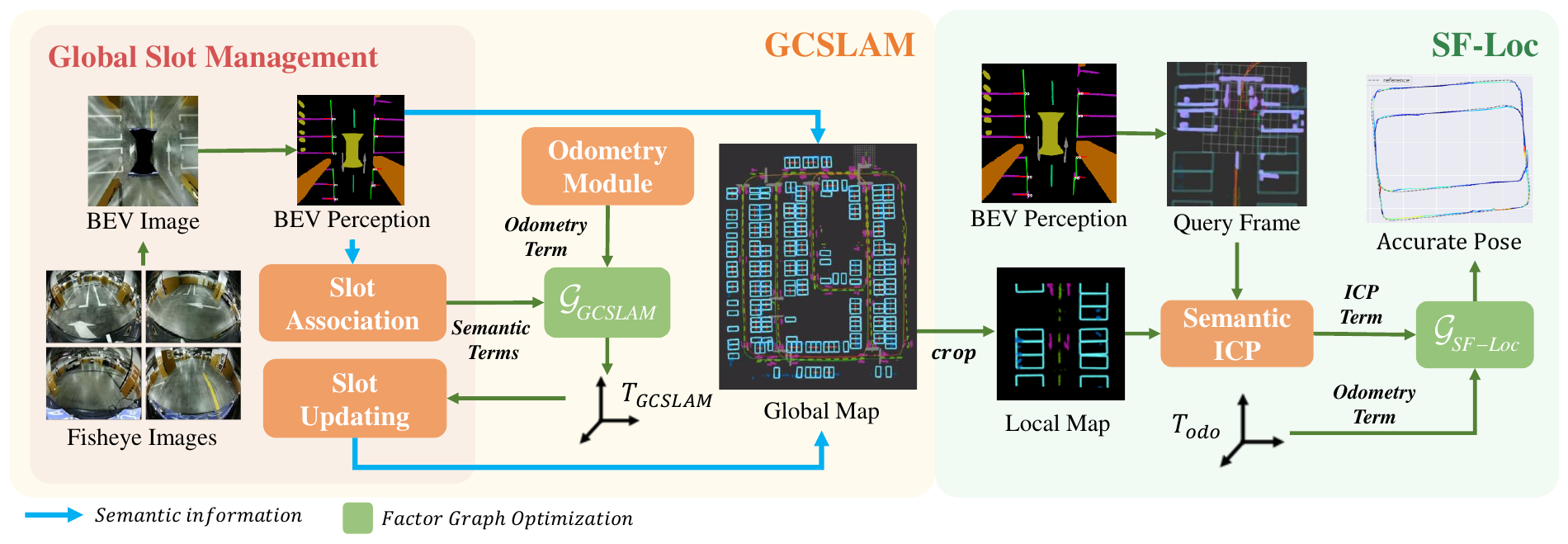}
    \vspace{-0.2in}
    \caption{Our system consists of three parts. The Global Slot Management module provides semantic information based on the BEV perception module. 
    By leveraging this semantic information, GCSLAM generates three semantic terms, including Registration Error Term, Adjacent Error Term, and Global Vertical Error Term. Subsequently, the semantic-constrained factor graph $\mathcal{G}_\text{GCSLAM}$ jointly optimizes the poses and semantic map to achieve a precise global map. 
    GCSLAM is only performed when entering a new parking lot for the first time. Once the map is created, SF-Loc provides precise localization results through a novel factor graph $\mathcal{G}_\text{SF-Loc}$.}
    \label{fig:frameWork}
    \vspace{-0.2in}
\end{figure*}

To address these issues, we propose a globally consistent semantic SLAM system GCSLAM, which consists of a semantic-constrained factor graph and a global slot management module designed for semantic mapping and robust tracking in parking lots.
Specifically, we introduce a new slot association strategy to accurately determine the relationships between parking slots.
Moreover, we design an adjacent error term and a global vertical error term in our semantic-constrained factor graph to mitigate the slot misalignment and distortion problem. The former error term is capable of reducing the gap between adjacent slots and the latter constrains parking slots to maintain correct orientations, which are caused by accumulative drift and noise detections.  
Furthermore, we introduce a global slot management module to store slot observations and update them accordingly. This module includes an unstable slot filter and a slot updater, which can address the false detections and noise problems caused by BEV perception module. 

Apart from GCSLAM, we propose a semantic-fusion localization subsystem (SF-Loc) for map-based localization. For an unknown parking lot, after constructing a complete global map using GCSLAM, we directly activate SF-Loc for map-based localization during subsequent revisits. Such subsystem typically provide higher accuracy and faster performance as it leverages prior scene knowledge established by GCSLAM. In SF-Loc, we design a novel factor graph that integrates semantic ICP registration and odometry for optimization. 
This design enables robust tracking in challenging scenarios, particularly addressing convergence failures in feature-sparse regions.

Our main contributions are as follows:
\begin{itemize}
    \item We propose a globally consistent semantic SLAM system GCSLAM, which is based on a semantic-constrained factor graph with novel semantic error terms for constraints, as well as innovative slot representation for joint optimization.
    \item We introduce a parking slot management module that stores slot observations and updates global slots, while also effectively handling noise and false detections.
    \item We propose a map-based localization subsystem SF-Loc, which integrates semantic-ICP registration and odometry constraints using factor graph optimization.
    \item We validate our system in complex real-world indoor parking lots, showing that our system achieves real-time, high-precision localization and semantic mapping performance.
    
\end{itemize}

\section{RELATED WORK}

Early visual SLAM approaches~\cite{davison2007monoslam, chiuso2002sfm, eade2006scalable} are implemented based on filtering methods. Subsequently, SLAM systems~\cite{strasdat2010scale, strasdat2011double, engel2017direct, mur2017orb2} utilizing bundle adjustment (BA) optimization emerged. DSO~\cite{engel2017direct} introduces both photometric error and geometric prior to estimate dense or semi-dense geometry. ORB-SLAM~\cite{mur2015orb} employs ORB features and sliding window to achieve precise pose estimation. Compared to filtering methods, optimization-based approaches offer higher accuracy and better global consistency.

Despite that, SLAM with a single camera is unable to recover scale and is vulnerable to visual ambiguities. To enhance the robustness and accuracy of the system, multi-sensor fusion methods~\cite{mourikis2007MSCKF, lee2020VIWO, qin2018vins} that combine visual data with other sensors are developed. MSCKF~\cite{mourikis2007MSCKF} constructs an observation model using visual information and updates the state with inertial measurement unit (IMU) data. VINS-Mono~\cite{qin2018vins} proposes a tightly-coupled, optimization-based visual-inertial system. VIWO~\cite{lee2020VIWO} develops a pose estimator based on MSCKF, which integrates IMU, camera, and wheel measurements. 
DM-VIO~\cite{von2022dm} enhances IMU initialization through delayed marginalization and pose graph bundle adjustment. Ground-Fusion~\cite{yin2024ground} introduces an adaptive initialization strategy to address multiple corner cases.

However, these methods are unable to perform SLAM within indoor parking slots for AVP tasks due to the complex conditions of indoor environments, such as limited distinctive features and complex lighting conditions.
To solve these problems, some works~\cite{qin2020avp, shao2020tightly, shao2021mofis, zhao2019visual} are proposed. 
These works all utilize bird's-eye view (BEV) images as input, which can provide rich ground features to tackle the issue of limited distinctive features in parking lots.
AVP-SLAM~\cite{qin2020avp} uses semantic segmentation to annotate parking spaces, ground markings, speed bumps, and other information in the images, as the segmentation method can effectively adapt to complex lighting conditions. 
This semantic information is added to a global map, which is then used for registration-assisted localization. However, their map is a pure point cloud map used for registration, without recording each parking lot independently and lacking important attribute information such as the location and angle of each parking slot. Zhao et al.~\cite{zhao2019visual}  utilize parking slot detector~\cite{li2017detect} to detect the entry points of parking slots and combine the observations of parking slots with odometry to construct new positioning factors. However, this method does not maintain an overall parking space map. Instead, it primarily uses the map as an auxiliary tool for localization. VISSLAM~\cite{shao2020tightly} adds constraints between parking spaces, combining odometry information to propose a parking slots management algorithm that improves mapping results. Subsequent work MOFISSLAM~\cite{shao2021mofis} incorporates sliding window optimization, achieving higher localization accuracy and improved mapping results.

Nevertheless, existing methods are sensitive to noise and perform poorly in complex parking lots. To address this, we propose a semantic-constrained factor graph for indoor parking SLAM, improving both robustness and accuracy.

\section{METHOD}
\label{section:Method}

Our system adopts multiple sensors as input, including a front-view camera, IMU, wheel encoder, and four surround-view cameras. The overall framework of our work is shown in Fig. \ref{fig:frameWork}. 
The first part of our work is the SLAM system GCSLAM. GCSLAM integrates three modules: odometry, global slot management, and semantic-constrained factor graph optimization. The odometry module is loosely coupled with other modules, making it replaceable and enhancing the system's flexibility and usability. In this paper, we employ VIW~\cite{ztd2021viwo} as our odometry. The global slot management module includes BEV perception module and slot association. Our BEV perception module is a multi-task framework based on \cite{peng2022pp, yolox, yolov8}. It takes BEV images as input to generate semantic segmentation results (ground markings) and slot detection results (parking boundary endpoints) in real-time, using a unified backbone network with different output heads for each specific task. 
Subsequently, the detection results are registered to global slots, followed by slot association process.
Moreover, based on the odometry poses, semantic information, and slot association results, our semantic-constrained factor graph jointly optimizes pose and global semantic map to achieve robust tracking and accurate semantic mapping.
After establishing a global semantic map, the second part of our work, localization subsystem SF-Loc, fuses odometry poses with semantic registration results for map-based localization.

\subsection{Semantic-constrained Factor Graph Optimization}
\label{section:System}
\begin{figure}[t]
    \centering
    \includegraphics[width=8cm]{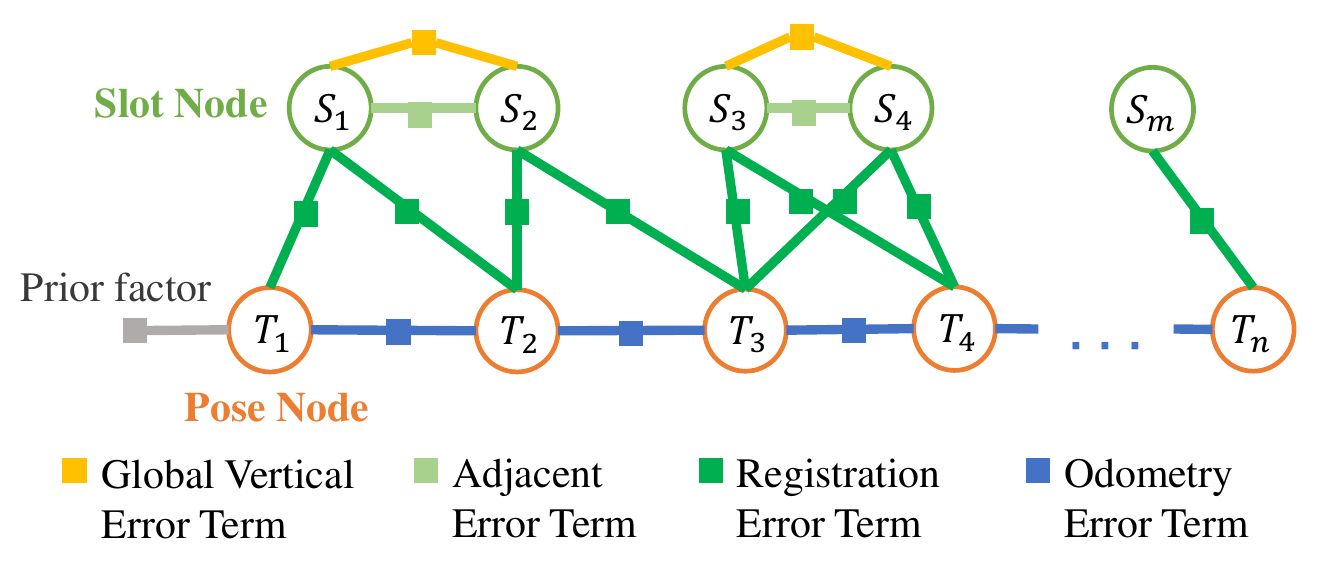}
    \vspace{-0.1in}
    \caption{The semantic-constrained factor graph structure of our SLAM system. The graph consists of two types of nodes: Pose Nodes and Slot Nodes. Four types of error terms are used to constrain the nodes. 
    The prior factor is provided by odometry and only applies to the first pose node, determining the absolute pose of the first pose.}
    \label{fig:Slam}
    \vspace{-0.2in}
\end{figure}

We view the SLAM task as a factor graph optimization problem, aiming to estimate the accurate pose of keyframes. The keyframe is selected based on the inter-frame distance provided by the odometry. A factor graph consists of nodes and edges, where nodes represent the variables to be optimized, and edges are error terms that constrain the nodes. As depicted in Fig. \ref{fig:Slam}, GCSLAM constructs the semantic-constrained factor graph $\mathcal{G}_\text{GCSLAM}$ using two types of nodes $\{T_{node}^i\}, \{S^k\}$ and four types of edges $\{E_{odo}^{i,j}\}, \{E_{reg}^{k,i}\}, \{E_{adj}^{k,p}\}, \{E_{vert}^{k,p}\}$:

\begin{equation}
    \small
    \mathcal{G}_\text{GCSLAM}=\{\{T_{node}^i\}, \{S^k\}, 
    \{E_{odo}^{i,j}\}, \{E_{reg}^{k,i}\}, \{E_{adj}^{k,p}\}, \{E_{vert}^{k,p}\}\}. 
    \mathrel{\normalsize}
\end{equation}
 
 The definitions of the nodes and error terms will be specifically introduced as follows.

\subsubsection{Pose Node} 
\label{sec:pose node}
The pose node $T_{node}^i$ stores 3 degree-of-freedom (DoF) car pose ($x,y,\theta$) of frame $i$ in world coordinates since our SLAM system assumes a planar parking lot. We initialize the pose node $T_{node}^i$ using estimated pose $T_{odo}^i$ provided by the odometry module. This module runs as a separate thread.

\subsubsection{Slot Node}
When BEV perception module detects a parking slot, it outputs the endpoint coordinates and direction of its entry edge in the pixel coordinate. We first register the midpoint of the entry edge to world coordinate using virtual intrinsic $K$ of BEV images and current frame pose $T^i$. The equation for $T^i$ is:
\begin{equation}
    T^i = T_{node}^{i-1} ({T_{odo}^{i-1}})^{-1}{T_{odo}^i}.
\end{equation}
Then, we perform slot association to determine the global ID $k$ of this observed slot. This global slot is denoted as $S_k$. We record the observation of $S_k$ in the current frame $i$ as $O_k^i$, representing the midpoint coordinate of $S_k$'s entry edge in the car coordinate of frame $i$. The 3DoF pose of $S_k$'s entry edge midpoint in the world coordinate is denoted as the slot node.

\subsubsection{Odometry Error Term (\textit{OET})}
\label{sec:Odometry Error Term}
We construct OET between $T_{node}^{i}$ and $T_{node}^{j}$ based on the odometry module. The specific form for OET is:
\begin{equation}
    E_{odo}^{i,j}=({T_{node}^i})^{-1} ({T_{node}^j)^{-1}} - ({T_{odo}^i})^{-1}T_{odo}^{j}.
\end{equation}

\subsubsection{Registration Error Term (\textit{RET})}
RET constrains the relationship between $S^k$ and $T_{node}^i$. By transforming the observation $O_k^i$ to the world coordinate through $T_{odo}^i$ and comparing it with $S^k$, we can establish RET as follows:

\begin{equation}
    E_{reg}^{k,i}={T^i}{O_k^i}-{S^k}.
\end{equation}

\subsubsection{Adjacent Error Term (\textit{AET})}
When the $i$-th keyframe arrives, all $O_k^i$ are traversed. If the distance $d_{k,p}=||O_k^i-O_p^i||$ between two slots $S_k$ and $S_p$ is less than a specified threshold (2.5m), they are considered adjacent. AET $E_{adj}^{k,p}$ is established between adjacent slots. We use this error term to ensure that the directions of adjacent slots are consistent and there are no gaps between them. Specific form for AET is:

\begin{equation}
\label{equation:e_adj}
    E_{adj}^{k,p}=\frac{\overrightarrow{w_k}+\overrightarrow{w_p}}{2}-[(S^k)_{1:2}-(S^p)_{1:2}],
\end{equation}
where $\overrightarrow{w_k}$ represents the entry edge of parking slot $S_k$, as shown in Fig. \ref{fig:adj-slots}. $(S^k)_{1:2}$ denotes the first two dimensions of $S^k$, representing the $xy$ coordinates of the slot.

\subsubsection{Global Vertical Error Term (\textit{GVET})}
In most parking lots, parking slots are either perpendicular or parallel to each other. To utilize this information, we introduce the new concept of the global slot direction $\overrightarrow{d_{global}}$. We define $\overrightarrow{d_{global}}$ as the average width $\overrightarrow{w}$ of the first five parking slots observed. This definition is chosen because SLAM is relatively accurate at the beginning, with no accumulated drift. GVET is only applied to adjacent slots, because there may be inclined slots that are not parallel to other slots in some parking lots. Such inclined parking slots are more likely to exist in isolation, so applying a global vertical constraint to these slots would be incorrect. The specific expression for GVET is:
\begin{equation}
    \begin{split}
    E_{vert}^{k,p}=min\{
        &|[(S^k)_{1:2}-(S^p)_{1:2}]\cdot \overrightarrow{d_{global}}|, 
        \\
        &|[(S^k)_{1:2}-(S^p)_{1:2}]\cdot \overrightarrow{d_{global}}^\perp|
    \}.
    \end{split}
\end{equation}

These error terms help maintain the accuracy and stability of GCSLAM. We observe that adding $E_{vert}$ significantly mitigates issues such as tilting and twisting during long-distance straight driving. The incorporation of $E_{adj}$ effectively corrects the accumulated drift of odometry. The related experiment is presented in Section \ref{Experiments}.

\subsection{Global Slot Management}
During the factor graph optimization process, the car poses are continuously optimized. After the car poses are optimized, the global slots' poses, registered from the car poses $T_i$ and slot observation $O_k^i$, need to be updated as well. Therefore, we use the Global Slot Management module to store and manage multiple frames of slot observations. When a new frame of observation arrives, the management system associates the current observation with an existing global slot. Otherwise, it creates a new global slot. The management system updates the global parking slots when the car's pose is optimized or when a new frame of observation arrives.

\begin{figure}[!htp]
    \centering
    \hspace{-5mm}
    \subfigure[Associate]{
        \label{fig:Associate}
        \includegraphics[height=1.88cm]{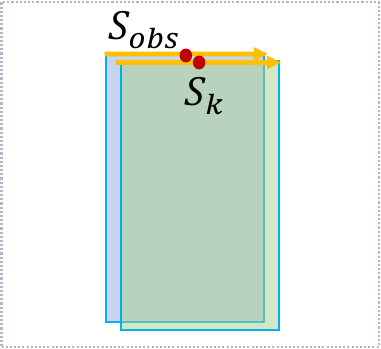}
    }
    \hspace{-5mm}
    \subfigure[Create]{
        \label{fig:Create}
        \includegraphics[height=1.88cm]{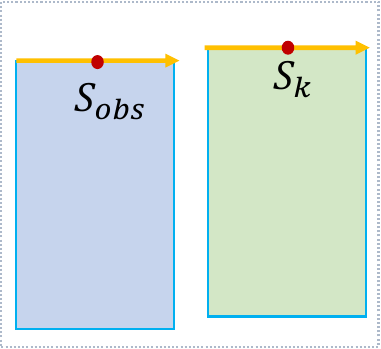}
    }
    \hspace{-5mm}
    \subfigure[Discard]{
        \label{fig:Discard}
        \includegraphics[height=1.88cm]{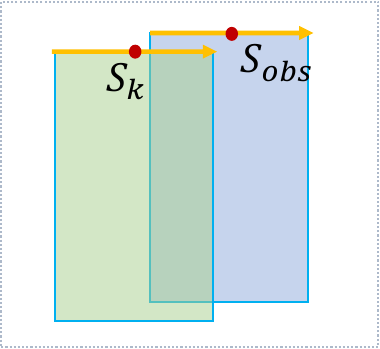}
    }
    \hspace{-5mm}
    \subfigure[Adjacent]{
        \label{fig:adj-slots}
        \includegraphics[height=1.88cm]{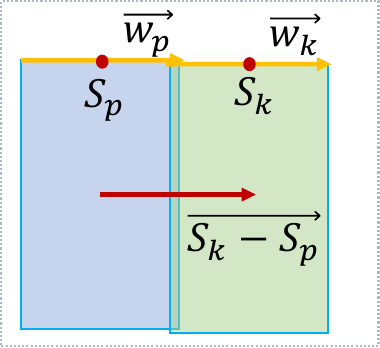}
    }
    \caption{Three possible situations when a new observation is made. Depending on the distance $d$ between the new observation and its nearest global slot, the new observation may be associated with an existing slot ($d\leq1$), created as a new global slot ($d\geq2$), or deemed a false detection and discarded ($1<d<2$).
    Specially, two slots will be regarded as adjacent if $2\leq d\leq2.5$.
    }
    \label{fig:association}
    \vspace{-0.1in}
\end{figure}

\subsubsection{Slot Association}
In order to determine whether a slot observation is associated with an existing global slot that is previously observed, we first register the current observation to the world coordinate, denoted as $S_{obs}$. We then use a kd-tree to find the nearest global slot. Based on the distance $d$ between their midpoints, we determine if they are associated. As shown in Fig. \ref{fig:association}, if $S_{obs}$ is not associated with any existing global slot, it will either be created as a new parking slot or discarded as a false detection. The specific parameters are shown in Fig. \ref{fig:association}.

Once a global slot $S_k$ is associated with $S_{obs}$, we increment the count of observation frames $c_O^k$ for this slot. By recording the observation frequency of each global slot, we can exclude low-frequency slots as false detections. This filter strategy can effectively mitigate the noise from the BEV perception module. The specific filter logic is illustrated in Alg. \ref{alg:filter}. When a new keyframe arrives, we perform slot association to all observed slots. Then, we iterate through all unstable global slots $S_k$. First, we increase $c_E^k$ (the frame count since $S_k$ is created,). If $S_k$ is observed in current frame, we increase $c_O^k$ (the frame count in which $S_k$ is observed). If $c_O^k>9$ we label $S_k$ as stable; otherwise, we continue to the next unstable slot. If an $S_k$ has $c_E^k>30$ and is still not labeled as stable, we consider it a false detection and delete the slot. These thresholds are selected based on comprehensive experimental data.

\begin{algorithm}[!h]
    \caption{Unstable Slots Filter}
    \label{alg:filter}

    \begin{algorithmic}[1] 
        \Statex \textbf{Input:}
        \State $c_E^k$: the frame count since $S_k$ is created;
        \State $c_O^k$: the frame count that $S_k$ is observed;
        \State $\mathcal{S}^i=\{S_k^i, S_p^i,...\}$: all slots observed in frame $i$;
        \State $\mathcal{S}_{un}=\{S_k, S_p, ...\}$: all unstable slots.
        \Statex \textbf{Output:} (Optional, specify if needed)
        
        \For{each $S_k \in \mathcal{S}_{un}$}
            \State $c_E^k \gets c_E^k + 1$;
            \If{$S_k \in \mathcal{S}^i$}
                \State $c_O^k \gets c_O^k + 1$;
            \EndIf

            \If{$c_O^k > 9$}
                \State label $S_k$ as stable;
            \ElsIf{$c_E^k > 30$}
                \State delete $S_k$;
            \EndIf
        \EndFor
    \end{algorithmic}
\end{algorithm}


\subsubsection{Slot Updating}

Since the factor graph is optimized in real-time, the car pose corresponding to each pose node is constantly changing. As the global slot is registered from the car pose $T^i$ and slot observation $O_k^i$, it should be updated accordingly when the pose of each frame in the factor graph changes. 

We assign a weight $w_k^i$ to each observation and consider multiple factors to calculate it: Detection Confidence $conf_{bev}$, Image Center Distance $dist_{IC}$, and Angle Weight $w_{rp}$. $conf_{bev}$ indicates the certainty of the perception result and is directly provided by the BEV perception module. Higher confidence means more reliable detection results. $dist_{IC}$ is the distance between the observed parking slot and the image center in pixel coordinates. A shorter image center distance implies that the parking slot is closer to the camera, resulting in clearer and more reliable observations. For areas where the plane assumption is violated, such as speed bumps, we design an angle weight $w_{rp}$. Larger roll and pitch angles indicate more severe violations of the plane assumption, making the current observation less reliable. Therefore, we design the angle weight as:
\begin{equation}
    w_{rp} =exp(-10\times\frac{\lvert{roll}\rvert+\lvert{pitch}\rvert}{2}).
\end{equation}

The overall weight for an observation is:
\begin{equation}
    w_k^i=0.2\times {conf}_{bev} + 0.5\times dist_{IC} + 0.3\times w_{rp}.
\end{equation}
The weighting factors in the equation were determined experimentally. We update the global slot and its weight by:
\begin{flalign}
\begin{split}
    &S_k'=\frac{S_k\times w_k \times (c_O^k-1) + S_{obs}\times w_k^i}
                {w_k \times (c_O^k-1) + w_k^i}, \\
    &w_k'=\frac{w_k \times (c_O^k-1) + w_k^i}
                {c_O^k}.
\end{split}
\end{flalign}

\subsection{Map-Based Localization Subsystem}
\label{SF-Loc}
\begin{figure}[t]
    \centering
    \includegraphics[width=8cm]{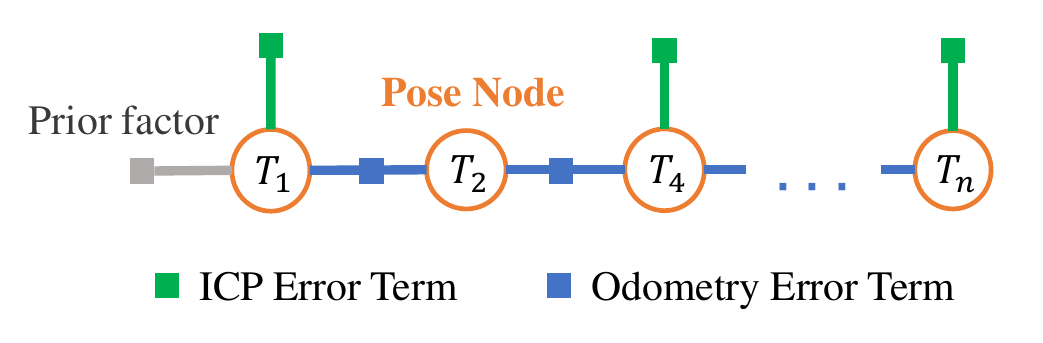}
    \vspace{-0.2in}
    \caption{Factor graph structure in SF-Loc. We use OET to constrain adjacent frames. Semantic ICP registration results are applied to the Pose Node through unary edges. Note that unary edges are not applied to every Pose Node.}
    \label{fig:match}
    \vspace{-0.2in}
\end{figure}

GCSLAM transforms the slots and other semantic information into point clouds and obtain a global map. 
Based on this global map, we propose a localization subsystem SF-Loc which fuses odometry pose and registration results. 
SF-Loc and GCSLAM are not activated simultaneously. GCSLAM is performed only during the first entry into an unknown parking lot, while SF-Loc is activated only when revisiting a parking lot with an established global map. 
As depicted in Fig. \ref{fig:match}, SF-Loc is constructed by a factor graph $\mathcal{G}_\text{SF-Loc}$:

\begin{equation}
    \mathcal{G}_\text{SF-Loc}=\{\{T_{node}^i\}, \{E_{odo}^{i,j}\}, \{E_{ICP}^{i}\}\}.
\end{equation}

There is only one type of pose node $\{T_{node}^i\}$, which is the same as the Pose Node introduced in Section \ref{sec:pose node}. The Odometry Error Term (\textit{OET}) is described in Section \ref{sec:Odometry Error Term}. Additionally, some nodes are constrained by Semantic ICP Error Terms, which are based on the semantic ICP registration.

Our semantic ICP registration algorithm performs registration between the local map and the current point cloud. The local map is a $30m\times30m$ map extracted from the global map based on the previous pose. The current point cloud is transformed from BEV semantic. During the semantic ICP process, the nearest neighbor with the same semantics for each point is identified using a kd-tree. Based on the matching relationship of semantic point pairs, the transformation between the current point cloud and the local map is calculated. This process is iteratively executed until convergence, providing a refined pose $T_{ICP}^i$. 

The Semantic ICP Error Terms are unary edges, providing the absolute pose result of the registration: 
\begin{equation}
    E_{ICP}^i=T_{ICP}^i-T_{node}^i.
\end{equation}

Due to the strong constraints imposed by unary edges and the high noise of semantic segmentation, we reduce the frequency of adding ICP unary edges. We add ICP Error Terms every 10 frames and perform jump detection before adding them. We calculate the distance between the current frame's ICP registration result and the previous frame's ICP registration result. If the distance exceeds a threshold of 2 meters, the current frame's registration result is considered inaccurate. In such cases, we do not add an ICP Error Term for the current frame.

The Semantic ICP Error Terms effectively correct the accumulated drift of the odometry, while OET mitigates the unstable jumps in the ICP. Thereby, SF-Loc enhances the precision and robustness of the localization.

\section{EXPERIMENTS}
\label{Experiments}
\begin{figure}[b]
    \vspace{-0.2in}
    \subfigure[Experimental platform]{
        \label{fig:test_car}
        \includegraphics[height=2.5cm]{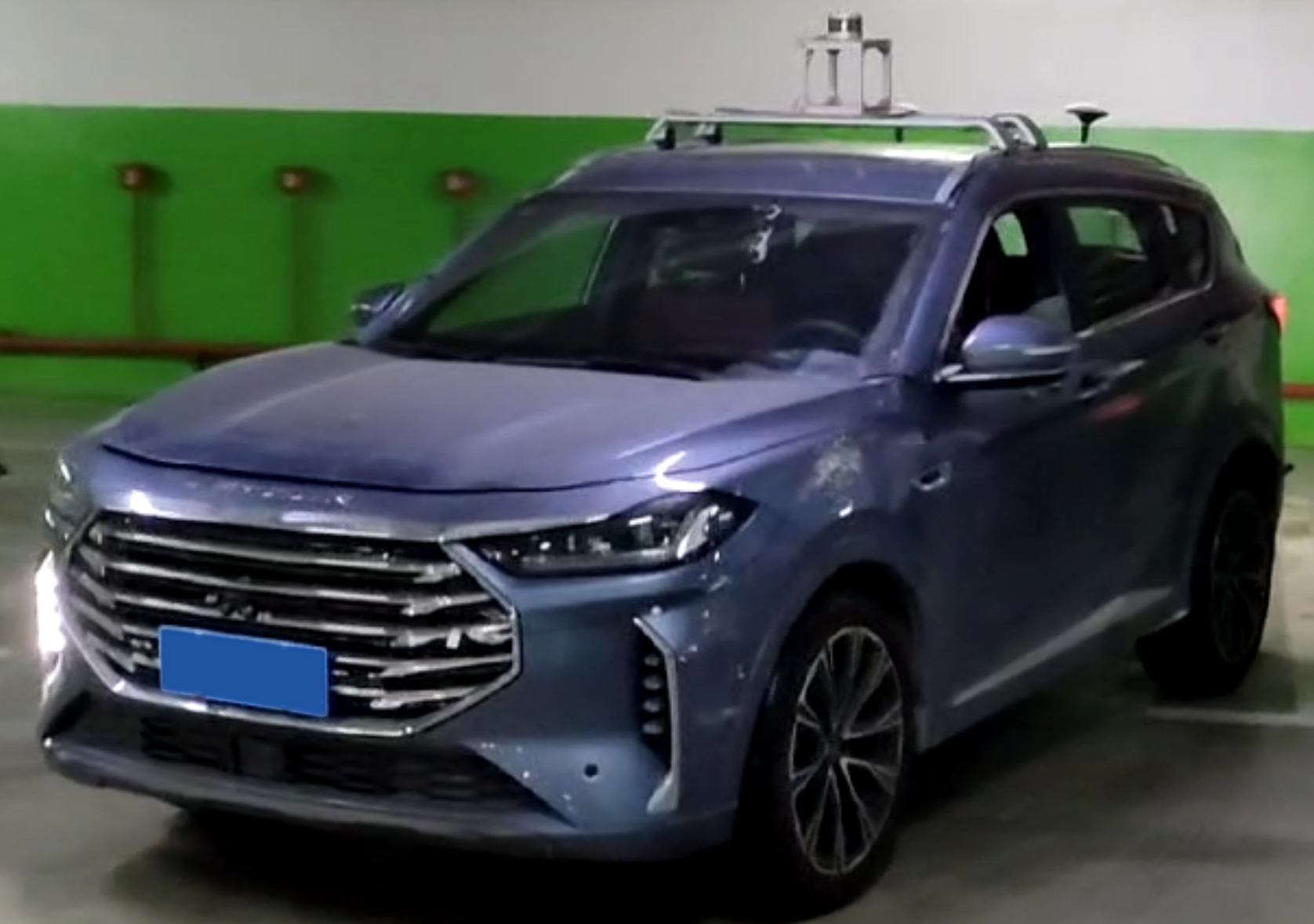}
    }
    \hspace{-3mm}
    \subfigure[Experimental environment]{
        \label{fig:front_img}
        \includegraphics[height=2.5cm]{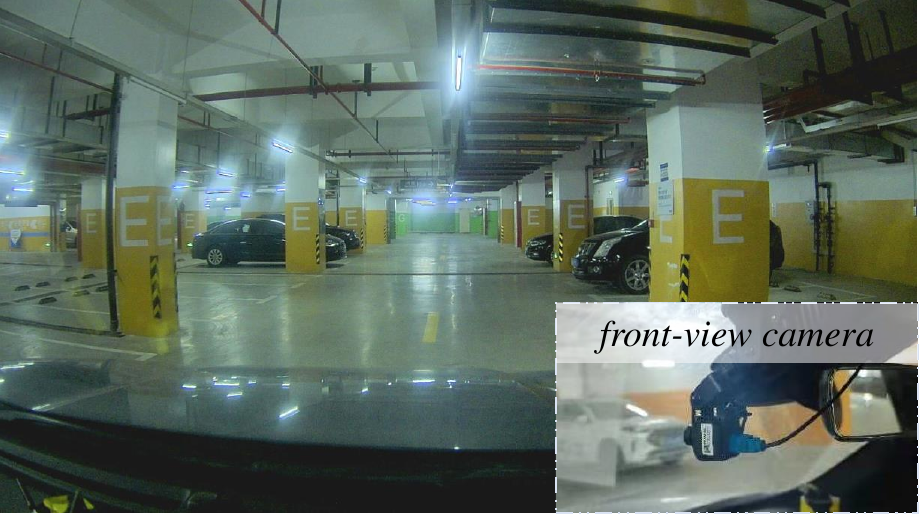}
    }
    \caption{The experimental platform and environment.}
    \label{fig:platform}
\end{figure}

We test our system in two underground parking lots. The environment of parking lots is shown in Fig. \ref{fig:front_img}. The sizes of both parking lots are $100m\times80m$ approximately. The test vehicle is shown in Fig. \ref{fig:test_car}, which is equipped with four surround-view fisheye cameras, one front-view camera, an IMU, wheel encoders, and a LiDAR. The LiDAR is only used to obtain the ground truth poses for evaluation. 
All sensors are calibrated offline. The IMU operates at 200\textit{Hz} and the wheel encoders run at 400\textit{Hz}. The front-view camera operates at 30\textit{Hz} with a resolution of $1920\times1080$ pixels. Each fisheye camera operates at 30\textit{Hz} with a resolution of $960\times540$ pixels. We conduct our experiments on an NVIDIA Jetson AGX Xavier.

\begin{figure}[t]
\hspace{-0.2in}
    \centering
    \includegraphics[width=8.5cm]{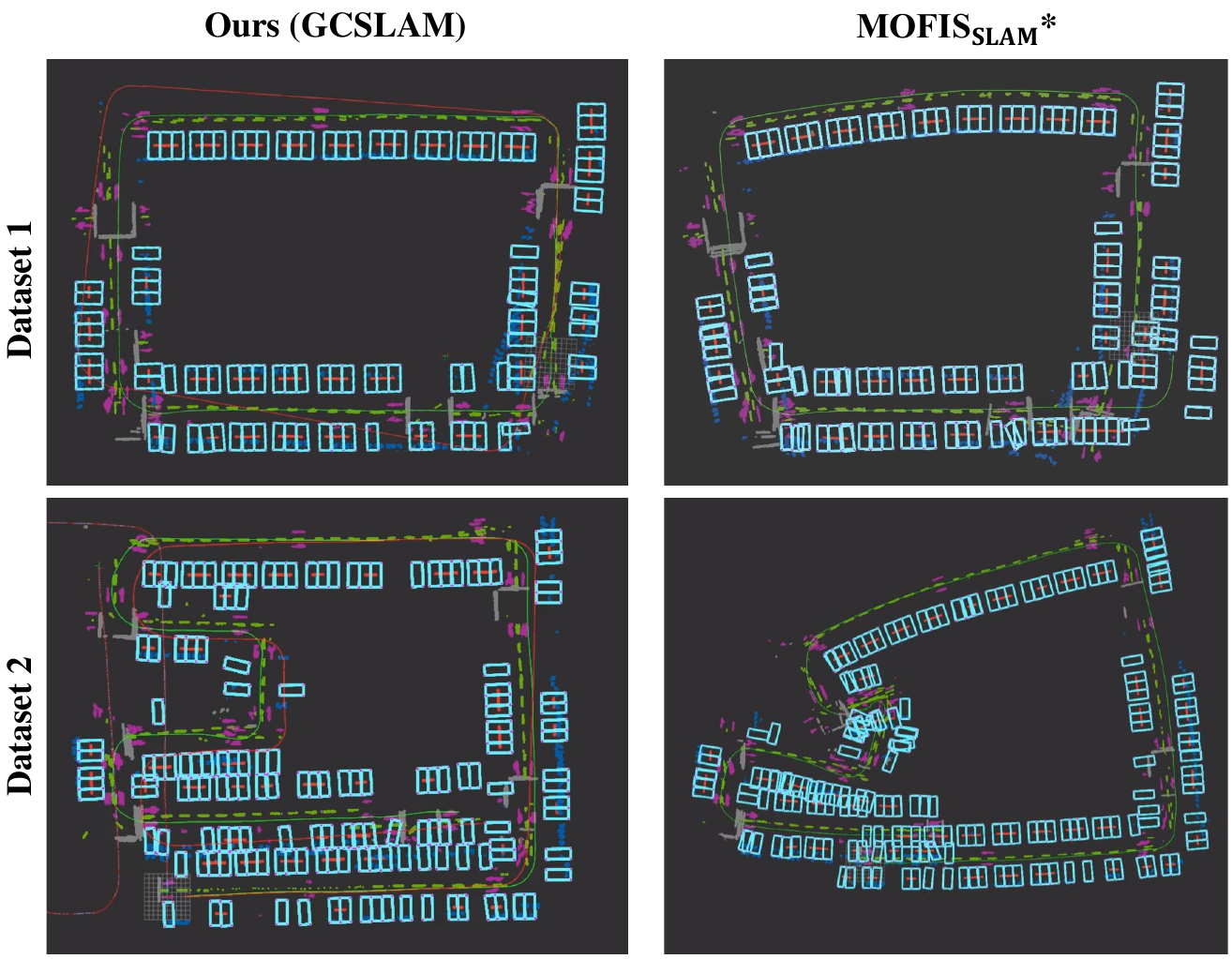}
      
    \vspace{-0.1in}
    \caption{The global maps constructed by GCSLAM and MOFISSLAM~\cite{shao2021mofis} on Dataset 1 and Dataset 2. $\text{MOFISSLAM}^*$~\cite{shao2021mofis} is a reproduction of our own implementation. For fairness, we incorporated wheel-related constraints into the process.}
    \vspace{-0.2in}
    \label{fig:global maps}
\end{figure}

Experiments are conducted on two representative datasets, which are collected by our test vehicle: \textit{(1)} In Dataset 1, the vehicle completes a square trajectory, returning to its starting point, covering a total distance of 379\textit{m}. \textit{(2)} In Dataset 2, the vehicle drives freely within the lot without returning to the origin, covering 438\textit{m}. 
The global maps built by GCSLAM on these datasets are shown in Fig. \ref{fig:global maps}.

\begin{table}[tb]
\centering
\caption{Absolute Trajectory Error in SLAM}
\vspace{-0.1in}
\resizebox{\linewidth}{!}{
\begin{tabular}{l|cc|cc}
\hline
  \multirow{2}{*}{Methods} & \multicolumn{2}{c|}{Dataset 1} & \multicolumn{2}{c}{Dataset 2}\\ 
  \cline{2-5}
                              & RMSE(m)~$\downarrow$  & NEES(\%)~$\downarrow$ & RMSE(m)~$\downarrow$  & NEES(\%)~$\downarrow$ \\ \hline
DM-VIO~\cite{von2022dm}       & \multicolumn{2}{c|}{fail}                 & \multicolumn{2}{c}{fail}    \\
ORB-SLAM3~\cite{campos2021orb}  & 13.130            & 3.462               & \multicolumn{2}{c}{fail}    \\
VIW~\cite{ztd2021viwo}        & 4.926               & 1.299               & 12.04               & 2.747 \\
Ground-Fusion~\cite{yin2024ground}& 4.342           & 1.145               & 6.387               & 1.457\\ 
Ours (GCSLAM)                         & \textbf{1.846}      & \textbf{0.487}      & \textbf{2.286}      & \textbf{0.522} \\ \hline
\end{tabular}}
\vspace{-0.1in}
\label{table:slam}
\end{table}

\subsection{Evaluation on GCSLAM}
\label{exp:slam}

\subsubsection{Tracking}
As existing SLAM algorithms for indoor parking~\cite{qin2020avp, shao2020tightly, shao2021mofis, zhao2019visual} are not open-source, and our experiments are conducted on AGX without sufficient computational resources for learning-based SLAM, we compare GCSLAM with some open-source traditional visual SLAM: ORB-SLAM2~\cite{mur2017orb2}, VIW~\cite{ztd2021viwo} DM-VIO~\cite{von2022dm} and Ground-Fusion~\cite{yin2024ground}. Due to the absence of GNSS data in the underground parking lot, we utilize a LiDAR odometry LeGO-LOAM~\cite{shan2018lego} to obtain the ground truth. 
We evaluate the ATE RMSE~\cite{sturm2012benchmark} and NEES (normalized estimation error squared) for each system. NEES is calculated as RMSE divided by total trajectory length. 

As shown in Table \ref{table:slam}, GCSLAM demonstrates significantly lower absolute trajectory error than ORB-SLAM3~\cite{campos2021orb}, VIW~\cite{ztd2021viwo} and Ground-Fusion~\cite{yin2024ground}. 
This is because our semantic-constrained factor graph effectively provides more stable and global semantic constraints, resulting in minimal cumulative drift, better accuracy and global consistency.
DM-VIO~\cite{von2022dm} fail due to the absence of wheel encoders. In indoor parking lots, cameras can not offer reliable data because of the complex lighting conditions, while wheel encoders can provide more accurate data, as the ground is flat and tire slippage is negligible.

Although most indoor parking SLAM are not open-source and cannot be compared with our method on the same dataset, the assessment of NEES can offer an insight into algorithmic accuracy.
Attributes to our novel error terms and modules, GCSLAM achieves a NEES of 0.487\%, which is significantly lower than the result of AVP-SLAM~\cite{qin2020avp} (1.33\%).

\begin{table}[t]
\centering
\caption{Mapping Error}
\vspace{-0.1in}
\resizebox{\linewidth}{!}{
\begin{tabular}{l|cc|cc}
\hline
  \multirow{2}{*}{Methods} & \multicolumn{2}{c|}{Dataset 1} & \multicolumn{2}{c}{Dataset 2}\\ 
  \cline{2-5}
                   & SWE(cm)~$\downarrow$  & AE(cm)~$\downarrow$ & SWE(cm)~$\downarrow$  & AE(cm)~$\downarrow$ \\ \hline
$\text{MOFISSLAM}^*$ \cite{shao2021mofis}    & 0.703                 & 4.224          & 1.729            & 4.279  \\
Ours (GCSLAM)      & \textbf{0.044}        & \textbf{2.146}   & \textbf{0.492}   & \textbf{0.776} \\ \hline
\end{tabular}}

\label{table:mapping}
\end{table}

\begin{table}[t]
\centering
\caption{Localization Error}
\vspace{-0.1in}
\resizebox{\linewidth}{!}{
\begin{tabular}{l|cc|cc}
\hline
  \multirow{2}{*}{Methods} & \multicolumn{2}{c|}{Dataset 1} & \multicolumn{2}{c}{Dataset 2}\\ 
  \cline{2-5}
                   & RMSE(m)~$\downarrow$  & NEES(\%)~$\downarrow$ & RMSE(m)~$\downarrow$  & NEES(\%)~$\downarrow$ \\ \hline
ICP                & 2.593                 & 0.688          & \multicolumn{2}{c}{fail}    \\
Semantic ICP       & 2.055                 & 0.545          & 2.124            & 0.498\\
Ours (SF-Loc)       & \textbf{2.013}      & \textbf{0.534}   & \textbf{1.925}   & \textbf{0.451} \\ \hline
\end{tabular}}
\vspace{-0.1in}
\label{table:reloc}
\end{table}

\begin{figure}[!htp]
    \centering
    \subfigure[]{
        \label{fig:adj_ablation}
        \includegraphics[height=2.9cm]{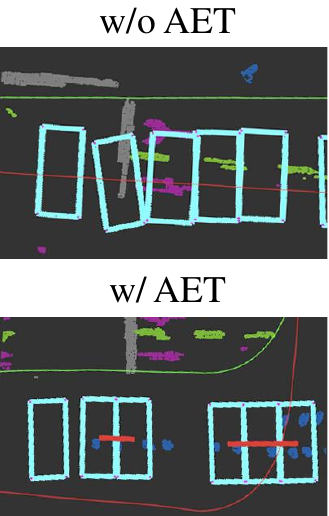}
    }
    \hspace{-3mm}
    \subfigure[]{
        \label{fig:filter_ablation}
        \includegraphics[height=2.9cm]{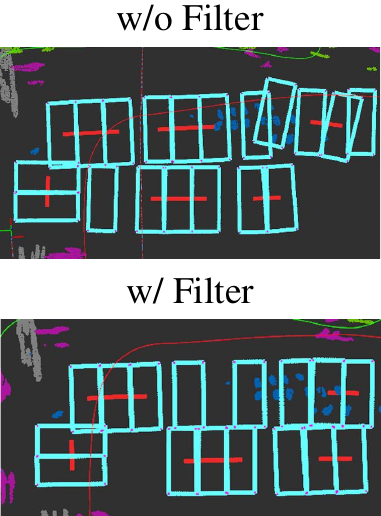}
    }
    \hspace{-3mm}
    \subfigure[]{
        \label{fig:vert_ablation}
        \includegraphics[height=2.9cm]{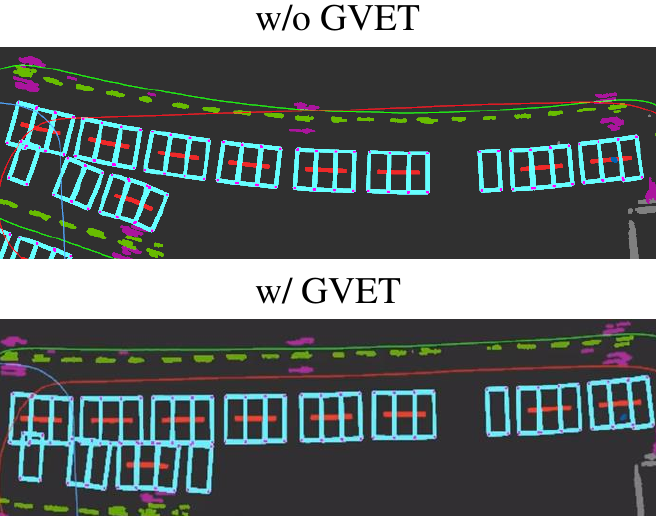}
    }
    \caption{The ablation study for (a) AET, (b) Unstable Slots Filter, and (c) GVET.}
    \vspace{-0.2in}
    \label{fig:ablation}
\end{figure}

\subsubsection{Mapping}
We evaluate mapping performance by two proposed metrics: Slot Width Error (\textit{SWE}) and Adjacent Error (\textit{AE}). SWE denotes the difference between the average parking slot width in the global map and the actual slot width, representing the gap between the global map and the real world. 
As shown in the Fig. \ref{fig:global maps}, we compared GCSLAM and $\text{MOFISSLAM}^*$~\cite{shao2021mofis} on our datasets. Table \ref{table:mapping} presents the accuracy results of both methods, demonstrating the precision of our algorithm. 
GVET improves the global consistency of GCSLAM, while the unstable slots filter and slot updater in global slot management module allow GCSLAM to better adapt to noise. 
Due to the absence of GVET and Global Slot Management, $\text{MOFISSLAM}^*$~\cite{shao2021mofis} exhibits misalignment and distortion in large-scale, complex parking lots.

\subsubsection{Ablation Study}
We also evaluate the effectiveness of different modules in mapping. The influence of different modules on mapping is shown in Fig. \ref{fig:ablation}. In Fig. \ref{fig:ablation}, the upper images display outcomes without different modules, while the lower images show results with them. As observed, removing AET results in irregularities between slots. GVET significantly mitigates the tilting issue during long-distance straight driving. Without the Unstable Slots Filter, the global map contains numerous erroneous slots.

\subsection{Evaluation on SF-Loc}
\label{exp:reloc}
As SF-Loc is based on known map localization, we focus on evaluating its accuracy within the global map. Therefore, We use the GCSLAM trajectory as the ground truth. 
The global map we use is shown in Fig. \ref{fig:global maps}. 
The Semantic ICP is our registration algorithm introduced in Section \ref{SF-Loc}. 
As shown in Table \ref{table:reloc}, vanilla ICP fails due to convergence failure in the sparse map area. Compared to vanilla ICP, semantic ICP significantly improved accuracy. Moreover, the SF-Loc outperformed semantic ICP results attributed to the semantic fusion factor graph.

\section{CONCLUSION}
In this paper, we present GCSLAM, a novel system for indoor parking tracking and mapping. GCSLAM incorporates a semantic-constrained factor graph and novel error terms, enabling robust and high-precision mapping in complex parking environments. Additionally, we develop a map-based localization subsystem SF-Loc. SF-Loc fuses registration results and odometry poses based on a novel factor graph, effectively enhancing localization precision. We validate our algorithm through real-world datasets, demonstrating the effectiveness and robustness of our system.

\addtolength{\textheight}{-11.5cm}   

\bibliographystyle{IEEEtran}  
\bibliography{IEEEabrv,root} 


\end{document}